\pdfoutput=1
\documentclass[10pt,twocolumn,letterpaper]{article}
\usepackage[pagenumbers]{cvpr}

\definecolor{cvprblue}{rgb}{0.21,0.49,0.74}
\usepackage{amsmath}
\usepackage{amssymb}
\usepackage{booktabs}
\usepackage{graphicx}
\usepackage{capt-of}
\usepackage{microtype}
\usepackage{xcolor}
\usepackage{xspace}
\usepackage{url}
\usepackage[breaklinks,colorlinks,citecolor=cvprblue,linkcolor=cvprblue,urlcolor=cvprblue]{hyperref}

\graphicspath{{figures/}}
\definecolor{cvprblue}{rgb}{0.21,0.49,0.74}

\newtoggle{codepublic}
\newcommand{\coccfg}{\textsc{CoC-CFG}\xspace}
\newcommand{\Afive}{Alpamayo~1.5\xspace}
\newcommand{\calA}{\mathcal{A}}

\newcommand{\CLBestCollAFAnchor}{0}
\newcommand{\CLBestCollAFCtl}{1}
\newcommand{\CLBestCollAFWOne}{0}
\newcommand{\CLBestCollAFZero}{0}
\newcommand{\CLBestDistGTAnchor}{0}
\newcommand{\CLBestDistGTCtl}{1}
\newcommand{\CLBestDistGTWOne}{0}
\newcommand{\CLBestDistGTZero}{0}
\newcommand{\CLBestPassAnchor}{0}
\newcommand{\CLBestPassCtl}{1}
\newcommand{\CLBestPassWOne}{0}
\newcommand{\CLBestPassZero}{0}
\newcommand{\CLBestScoreAnchor}{0}
\newcommand{\CLBestScoreCtl}{1}
\newcommand{\CLBestScoreWOne}{0}
\newcommand{\CLBestScoreZero}{0}
\newcommand{\CLCollAFAnchor}{10}
\newcommand{\CLCollAFCtl}{7}
\newcommand{\CLCollAFWOne}{13}
\newcommand{\CLCollAFZero}{21}
\newcommand{\CLDistGTAnchor}{1.272}
\newcommand{\CLDistGTCtl}{1.118}
\newcommand{\CLDistGTWOne}{1.232}
\newcommand{\CLDistGTZero}{1.239}
\newcommand{\CLPassAnchor}{254}
\newcommand{\CLPassCtl}{262}
\newcommand{\CLPassWOne}{252}
\newcommand{\CLPassZero}{243}
\newcommand{\CLScenes}{300}
\newcommand{\CLScoreAnchor}{0.691}
\newcommand{\CLScoreCtl}{0.712}
\newcommand{\CLScoreWOne}{0.695}
\newcommand{\CLScoreZero}{0.638}
\newcommand{\DiffusionSteps}{10}
\newcommand{\NuRecAnchorW}{0.5}
\newcommand{\NuRecBestFixedW}{0.5}
\newcommand{\NuRecEvalScenes}{358}
\newcommand{\NuRecPeakGainPct}{16.5}
\newcommand{\NuRecPoolScenes}{120}
\newcommand{\OffBestPublished}{8.079}
\newcommand{\OffBestPublishedName}{DriveMA-4B}
\newcommand{\OffQTwo}{8.135159}
\newcommand{\OneSessBestW}{0.75}
\newcommand{\OneSessFixedTotal}{7}
\newcommand{\OneSessQTwo}{8.3032}
\newcommand{\OneSessQTwoOverFixedOne}{+0.0669}
\newcommand{\OneSessQTwoOverFixedThreeQuarter}{+0.0620}
\newcommand{\PerStepOverSceneLevel}{+0.0867}
\newcommand{\PerStepScore}{8.1972}
\newcommand{\ShroomBestIoU}{0.51}
\newcommand{\WaymoEvalFrames}{358}
\newcommand{\WaymoTrainFrames}{120}
\newcommand{\WaymoValFrames}{478}

\newcommand{\teasercaption}{%
\textbf{Not whether to trust a VLA's reasoning, but how strongly it steers the
action, at every step.} A reasoning VLA writes its reasoning, then decodes the
action conditioned on it. (a) \coccfg{} sets with one weight per decoding (Euler)
step how strongly the reasoning steers, from removed ($0$) through the original
model ($1$) to amplified ($2$). Held at every step, a larger weight turns the plan
of (b) harder and carries the plan of (c) farther. (b) On a held-out scene of the
Alpasim driving simulator the reasoning calls for a lane change the scene does not
offer; the original model (red) follows it off the recorded path (green). (c) This
reasoning reads correctly, yet followed at full strength the plan runs far ahead
of the recorded drive. In both, ours (blue; Q-adaptive guidance trained in
Alpasim) attenuates the reasoning.%
}

\newcommand{\papertitlebare}{How Much Should a VLA Trust Its Reasoning?\\
Step-Level Guidance as a Measured Action-Value Function}

\newcommand{\papertitle}{\papertitlebare}
\title{\papertitle}

\author{Dabin Seo \qquad Eunhyek Joa$^{*}$\\
Department of Mechanical Engineering, Seoul National University\\
{\tt\small sdabin@snu.ac.kr \qquad eunhyekj@snu.ac.kr}
}

\hypersetup{pdftitle={How Much Should a VLA Trust Its Reasoning? Step-Level Guidance as a Measured Action-Value Function},
  pdfauthor={Dabin Seo, Eunhyek Joa}}

\begin{document}

\twocolumn[{%
\renewcommand\twocolumn[1][]{#1}%
\maketitle
\begin{center}
  \centering
  \includegraphics[width=\textwidth]{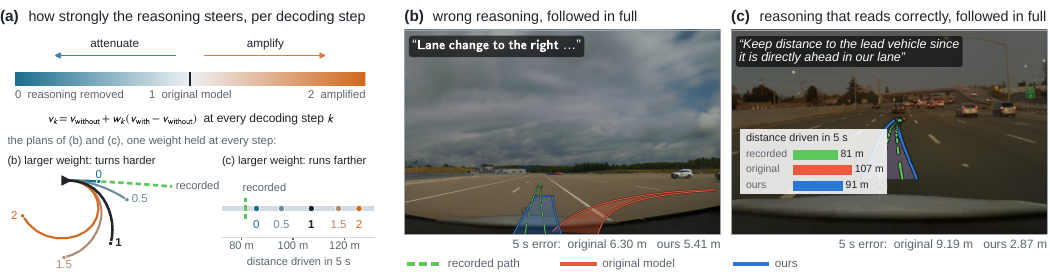}
  \captionof{figure}{\teasercaption}
  \label{fig:teaser}
  \vspace{0.3em}
\end{center}%
}]
{\renewcommand{\thefootnote}{$*$}\footnotetext{Corresponding author.}}

\begin{abstract}
Reasoning vision--language--action (VLA) policies write a reasoning trace and
then decode the action from it. A wrong trace yields a wrong action, and even a
correct-looking trace can mislead the plan. Existing remedies ask whether the
reasoning is right. We ask how strongly it should steer.
\textbf{CoC-CFG} applies classifier-free guidance between the policy's context
and the same context without the reasoning span. One weight per Euler step
attenuates or amplifies the reasoning, and $w{=}1$ recovers the original policy.
We treat the choice of this weight as a Markov decision process whose action values
are measured, not bootstrapped. With the expert frozen and the noise fixed, applying a
weight for one step and finishing the plan under the current controller returns
that weight's value exactly. \textbf{Q-adaptive guidance} fits these values with
small critics and improves the controller conservatively. In the Alpasim
driving simulator it outperforms every fixed weight and online RL (SAC, GRPO) in open loop,
and in closed loop it passes \CLPassCtl{} of \CLScenes{} scenes, against
\CLPassWOne{} for the original policy. On held-out frames of the Waymo
end-to-end driving benchmark, its first three stages outperform every fixed
weight, four adaptive-guidance methods, SAC and GRPO. On the hidden test set,
a frozen Alpamayo~1.5 with rule-based camera and vehicle conversions and
Q-adaptive guidance sets a new state of the art among published methods:
\OffQTwo{} rater feedback score, against \OffBestPublished{} for the
best published method.

\end{abstract}

\section{Introduction}
\label{sec:intro}

Vision--language--action (VLA) models increasingly reason before they act: a
vision--language model (VLM) writes a reasoning trace, and an action expert
decodes the action conditioned on it
\cite{nvidia2025alpamayo,zhou2025autovla,huang2026mindvlau1}. In the Alpamayo
driving planners \cite{nvidia2025alpamayo}, the VLM writes a Chain-of-Causation
(CoC) into its key--value cache, and a flow-matching expert decodes the
trajectory from that cache in a few Euler steps. The trace is not a caption but
part of the condition: editing it moves the trajectory
\cite{nguyen2026vladrivebench}.

The trace is often wrong. Fewer than half of a driving VLA's traces match the
scene \cite{mayumu2026faithful}, VLMs answer driving questions from the text
rather than the image \cite{xie2025drivebench}, and hallucination has
statistical causes \cite{kalai2025why}. A wrong trace becomes a wrong plan: in
Fig.~\ref{fig:teaser}b the trace calls for a lane change the scene does not
offer, and the unmodified policy follows it off the logged path. A trace that
reads correctly can mislead too: followed in full, the plan of
Fig.~\ref{fig:teaser}c runs far past the log.

The obvious fixes ask whether the reasoning is right. \Afive is already
post-trained with reinforcement learning for reasoning--action consistency
\cite{nvidia2025alpamayo}, and its traces still err; detecting hallucinations
remains unsolved, at \ShroomBestIoU{} IoU for the best system of a recent
shared task \cite{vazquez2026shroom}, and a detector only keeps or drops the
trace. Neither helps when the trace reads correctly yet misleads the plan.

We ask instead how strongly the trace should steer. \coccfg{},
Chain-of-Causation classifier-free guidance \cite{ho2022classifierfree}, runs
the frozen expert on the policy's own context and on the same context with only
the reasoning span removed, and moves each Euler step along the line through the
two velocities. The position on that line, the guidance
weight $w_k$, sets how strongly the trace steers step $k$: $w_k{=}1$ is the
unmodified model, lower weights attenuate the trace and higher weights amplify it
(Fig.~\ref{fig:teaser}a). Trusting and rejecting the trace are two points on this
axis.

\begin{figure}[t]
\centering
\includegraphics[width=0.84\columnwidth]{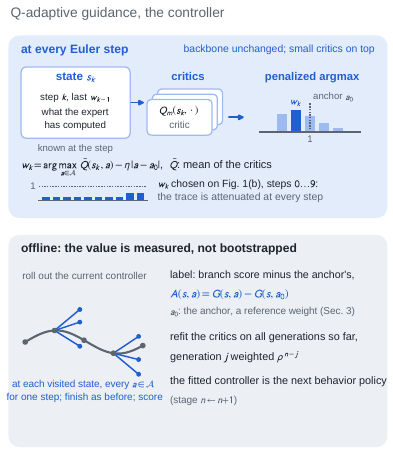}
\caption{\textbf{Q-adaptive guidance.} At every Euler step small critics read
what the step has computed, and the penalized argmax of their mean is $w_k$.
Offline, each weight's value is measured by one-step branches, and the critics
are refit on every generation of labels.}
\label{fig:method}
\end{figure}

The weight acts before the plan exists, so we treat its choice as a Markov
decision process over the Euler steps whose only reward is the score of the
finished plan (Sec.~\ref{sec:method}). Because the expert is frozen and the noise fixed,
generation is deterministic and can be reset to any step: applying one weight
for one step and finishing the plan under the current controller returns that
weight's value exactly. Learning becomes supervised regression on measured
values, where online reinforcement learning would bootstrap what can be
measured. \emph{Q-adaptive guidance}
(Fig.~\ref{fig:method}) fits small critics to these measurements with a cost
for leaving a reference weight, re-measures under the improved controller and
refits on every measurement so far.

In Alpasim, a driving simulator reconstructed from real drives, the controller
lowers the open-loop displacement cost of \Afive by \NuRecPeakGainPct\,\%
against $w{=}1$, above soft actor-critic (SAC) and group-relative policy
optimization (GRPO) trained on the same signal, and passes \CLPassCtl{} of
\CLScenes{} closed-loop scenes against \CLPassWOne{} (Sec.~\ref{sec:alpasim}).
On the Waymo end-to-end driving benchmark \cite{xu2026wode2e}, its first three
stages clear every fixed weight, four adaptive-guidance methods, SAC and GRPO
on the held-out frames, and \Afive,
converted to the Waymo vehicle by rule-based conversions and steered by Q-adaptive
guidance, sets a new state of the art among published methods on the hidden
test set: \OffQTwo{} rater feedback score (RFS) against
\OffBestPublished{} (Sec.~\ref{sec:waymo}).

Our contributions are three.
(i) \textbf{\coccfg{}}, a guidance axis over the reasoning span of a VLA's
context that attenuates or amplifies the trace and is exact at $w{=}1$: an
unreliable trace becomes an influence to control rather than a defect to detect.
(ii) \textbf{Step-level guidance as a Markov decision process with measured
action values}, each obtained by a one-step branch, and \textbf{Q-adaptive
guidance}, a controller that chooses the weight at every Euler step from critics
fitted to these values.
(iii) \textbf{Evidence on a frozen driving VLA}: open-loop gains in Alpasim
above fixed weights and online RL, closed-loop gains over fixed weights,
held-out Waymo gains over fixed weights, adaptive-guidance methods and online RL, and
a new state of the art among published methods on the hidden test set of the
Waymo end-to-end driving benchmark.

\section{Related Work}
\label{sec:related}

\paragraph{Guidance}
Classifier-free guidance \cite{ho2022classifierfree} carries over to flow matching
\cite{lipman2023flow}; schedules \cite{kynkaanniemi2024interval} and learned or
searched per-step weights
\cite{galashov2025learntoguide,nevo2026pathguide,papalampidi2025dynamicguidance}
assume a trusted condition, whereas ours is the measured value of acting on a
condition that can be wrong.

\paragraph{Sampling as control, and offline value learning}
Denoising is a Markov decision process with a terminal reward \cite{black2023ddpo}, steered by the
gradient of a soft value \cite{uehara2024finetuning} or by soft value-based
selection among candidates \cite{li2024svdd}; our controller is the discrete,
one-dimensional member of this family (Sec.~\ref{sec:coccfg}), with a measured value. The fit is a
one-step advantage \cite{brandfonbrener2021onestep} (Sec.~\ref{sec:method}), conservatism comes from a
cost for departing from a reference weight, as in conservative policy iteration
\cite{kakade2002approximately}, and
the one-step branches are ``vine'' rollouts \cite{schulman2015trpo}.
\section{CoC-CFG: How Strongly the Reasoning Steers}
\label{sec:coccfg}

\paragraph{Reasoning VLAs with a flow-matching expert}
A reasoning VLA acts in two stages. A VLM reads the
observation and the instruction and writes a reasoning trace. An action expert
then decodes the action from a context $C$ that holds the observation, the
instruction and the trace; in the Alpamayo planners $C$ is the VLM's key--value
cache. The expert is a flow-matching model: from initial noise $x_0$ it
integrates a learned velocity field $v(x, t, C)$ over unit pseudo-time in $K$
Euler steps, $x_{k+1} = x_k + \Delta t\, v(x_k, t_k, C)$ with $\Delta t = 1/K$
and $t_k = k\,\Delta t$, and a decoder maps $x_K$ to the plan. A task score
$R(x_K)$, higher is better, rates the decoded plan against a reference the
policy does not observe, the logged future in driving. The tokens of the trace
occupy one contiguous span of $C$, the \emph{reasoning span}.

\paragraph{The axis}
Let $C_{\text{full}}$ be the complete context and $C_{\text{ref}}$ the same
context with the reasoning span removed and nothing else changed: observation,
instruction, prompt format and the positions of the remaining entries are kept.
\coccfg{} is classifier-free guidance \cite{ho2022classifierfree} between these
two branches. With a weight $w_k$ at step $k$ the Euler step becomes
\begin{equation}
\begin{split}
  x_{k+1} = x_k + \Delta t \Big( &v(x_k,t_k,C_{\text{ref}}) \\
  + w_k \big[ &v(x_k,t_k,C_{\text{full}}) - v(x_k,t_k,C_{\text{ref}}) \big] \Big),
\end{split}
  \label{eq:coccfg}
\end{equation}
$v_{\mathrm{with}} = v(x_k,t_k,C_{\text{full}})$ and $v_{\mathrm{without}} = v(x_k,t_k,C_{\text{ref}})$ denote
the two velocities (Fig.~\ref{fig:teaser}a). At $w_k{=}1$ the policy is the
unmodified VLA, at $w_k{=}0$ the trace is removed, weights in between attenuate
it and weights above $1$ amplify it. The weight takes values in a finite
\emph{action set} $\calA \ni 1$, fixed per backbone. The \emph{anchor}
$a_0 \in \calA$ is a reference constant weight, the best one on the training
scenes (Sec.~\ref{sec:alpasim}) or, on Waymo, the backbone's default $w{=}1$;
the controller's departures from it
are penalized (Sec.~\ref{sec:method}).

\section{Q-Adaptive Guidance}
\label{sec:method}

\paragraph{Step-level guidance as a Markov decision process}
A \emph{guidance rule} $\pi$ chooses the weight $w_k \in \calA$ at every Euler
step from what the expert has computed up to that step. Its \emph{state} is
$s_k = (k, x_k, w_{k-1}, C)$, with $w_{-1} := a_0$; the two velocities are
functions of $(x_k, t_k, C)$ and so of $s_k$. Its \emph{action} is the weight.
The transition is the guided Euler step \eqref{eq:coccfg}, deterministic for a
fixed initial noise, and the reward is zero until the last step and then the
score $R(x_K)$ of the decoded plan. The rule is non-anticipative: it reads
neither the scoring reference nor the score. Since the next latent depends on the
past only through $s_k$, this is a finite-horizon Markov decision process with
a terminal reward, and the best rule applies at every step the weight with the
highest \emph{action value} $Q(s_k, a)$, the expected final score of applying
$a$ at $s_k$ and choosing well afterwards.

\paragraph{The action value is measured, not bootstrapped}
Generation is \emph{deterministic}: the expert is frozen and the initial noise
fixed, so the same state and weights give the same plan. It is
\emph{resettable}: a stored latent restores the generation at any step. On a
training scene, where the scoring reference is known, the value of applying $a$
at $s_k$ and then following a fixed controller $\pi$ is therefore measured, not
estimated: a \emph{one-step branch} applies $a$ for one step, finishes the plan
under $\pi$, and its score $G^\pi(s_k, a)$ is that value with no sampling
variance. Paired with the anchor's branch from the same state, the \emph{label}
$A(s_k, a) = G^\pi(s_k, a) - G^\pi(s_k, a_0)$ isolates the effect of one step's
weight. Every weight is labeled at every visited state, and no target is built
from a later estimate: no bootstrapping, no $\max$ over next actions. Learning
the controller is supervised regression on measured values. Online
reinforcement learning fits this structure poorly: a group-relative policy
gradient \cite{shao2024deepseekmath} credits one score to a whole schedule of
$K$ decisions, and a soft actor-critic \cite{haarnoja2018sac} bootstraps a
critic toward values that can be measured exactly (Sec.~\ref{sec:alpasim}
compares the three).

\paragraph{Measuring}
Stage $n$ captures a generation of labels $D_n$ on the training scenes. Each
scene is rolled forward under the current controller $\pi_{n-1}$, with
$\pi_0 \equiv a_0$, with occasional non-anchor weights for exploration. At
every visited state all
$|\calA|$ weights are branched for one step, and each branch is completed under
$\pi_{n-1}$ and scored.

\paragraph{Fitting}
The critics regress a target that charges for leaving the anchor,
\begin{equation}
\begin{split}
  y(s,a) = \mathrm{clip}\big( &A(s,a) - \lambda_{c}\,\mathbf{1}[a \neq a_0] \\
  &- \lambda_{m}\,|a - a_0|,\; -\tau,\; \tau \big),
\end{split}
  \label{eq:target}
\end{equation}
a fixed cost $\lambda_c$ for any departure, a proportional cost $\lambda_m$ for
its distance, and a clip $\tau$ for the heavy tail of the score. A controller
fitted to these labels visits states its predecessor did not, where no label
was measured; the costs keep each improvement close to the measured behavior,
as in conservative policy iteration \cite{kakade2002approximately}.

\paragraph{Iterating}
The fitted controller $\pi_n$ becomes the next behavior policy, and the capture
repeats on the same training scenes. Labels age, since $D_j$ was measured under
the continuation of $\pi_{j-1}$, so stage $n$ fits every generation so far,
discounted in age:
\begin{equation}
  \mathcal{L}_n \;=\; \sum_{j=1}^{n} \rho^{\,n-j}\,\mathcal{L}_{D_j},
  \qquad 0 < \rho \le 1,
  \label{eq:replay}
\end{equation}
with $\mathcal{L}_{D_j}$ the loss of each critic $Q_m$ on $D_j$.

\paragraph{Deployment}
At inference the controller applies
\begin{equation}
  w_k = \arg\max_{a \in \calA}
  \big( \bar{Q}(s_k, a) - \eta\,|a - a_0| \big),
  \label{eq:selector}
\end{equation}
with $\bar{Q}$ the mean of the critics $Q_m$. The critics read features of
$s_k$, computed from the context and from what the denoising loop has computed;
no scoring reference and no branch rollout is used at inference.

\section{Experiments}
\label{sec:experiments}

\subsection{Alpasim: Open and Closed Loop}
\label{sec:alpasim}

\begin{table}[t]
\caption{\textbf{Alpasim open loop, held-out, \Afive.} Mean reward (negative
displacement cost) over \NuRecEvalScenes{} scenes, and the cost reduction
against the default $w{=}1$. SAC and GRPO at their best held-out checkpoint.
\textbf{Bold}: best reward.}
\label{tab:nurec}
\centering
\footnotesize
\setlength{\tabcolsep}{4pt}
\begin{tabular}{@{}lrr@{}}
\toprule
 & reward $\uparrow$ & vs.\ $w{=}1$ \\
\midrule
$w{=}0$ (span removed) & $-1.1315$ & $+3.3$\,\% \\
$w{=}1$ (default) & $-1.1704$ & -- \\
best fixed $w{=}\NuRecBestFixedW{}$ (anchor) & $-1.0495$ & $+10.3$\,\% \\
SAC, best checkpoint & $-1.0468$ & $+10.6$\,\% \\
GRPO, best checkpoint & $-1.0091$ & $+13.8$\,\% \\
\midrule
ours, stage $n{=}1$ & $-0.9943$ & $+15.0$\,\% \\
ours, stage $n{=}2$ & $-0.9802$ & $+16.3$\,\% \\
ours, stage $n{=}3$ (driven in Table 2) & $\mathbf{-0.9767}$ & $+16.5$\,\% \\
ours, stage $n{=}4$ & $-0.9769$ & $+16.5$\,\% \\
\bottomrule
\end{tabular}

\end{table}

\begin{table}[t]
\caption{\textbf{Alpasim closed loop, \Afive.} \CLScenes{} scenes outside the
training pool, 10\,s at 10\,Hz inside the reconstruction after ground-truth
warm-up, one inference per 25 control steps, one rollout per scene and arm;
ours is the stage-$n{=}3$ controller. A rollout passes when it ends without an
at-fault (front or lateral) collision or an off-road event; distance to log is
the largest gap to the logged path, averaged over scenes; score is the
simulator's composite rollout score. \textbf{Bold}: best in a column.}
\label{tab:closedloop}
\centering
\footnotesize
\providecommand{\clb}[2]{\ifnum#1=1 \textbf{#2}\else #2\fi}
\setlength{\tabcolsep}{3pt}
\begin{tabular}{@{}lcccc@{}}
\toprule
 & passed & at-fault coll. & dist.\ to log (m) $\downarrow$ & score $\uparrow$ \\
\midrule
$w{=}0$ (span removed) & \clb{\CLBestPassZero}{\CLPassZero{}/\CLScenes{}} & \clb{\CLBestCollAFZero}{\CLCollAFZero{}} & \clb{\CLBestDistGTZero}{\CLDistGTZero{}} & \clb{\CLBestScoreZero}{\CLScoreZero{}} \\
$w{=}\NuRecAnchorW{}$ (anchor) & \clb{\CLBestPassAnchor}{\CLPassAnchor{}/\CLScenes{}} & \clb{\CLBestCollAFAnchor}{\CLCollAFAnchor{}} & \clb{\CLBestDistGTAnchor}{\CLDistGTAnchor{}} & \clb{\CLBestScoreAnchor}{\CLScoreAnchor{}} \\
$w{=}1$ (default) & \clb{\CLBestPassWOne}{\CLPassWOne{}/\CLScenes{}} & \clb{\CLBestCollAFWOne}{\CLCollAFWOne{}} & \clb{\CLBestDistGTWOne}{\CLDistGTWOne{}} & \clb{\CLBestScoreWOne}{\CLScoreWOne{}} \\
ours & \clb{\CLBestPassCtl}{\CLPassCtl{}/\CLScenes{}} & \clb{\CLBestCollAFCtl}{\CLCollAFCtl{}} & \clb{\CLBestDistGTCtl}{\CLDistGTCtl{}} & \clb{\CLBestScoreCtl}{\CLScoreCtl{}} \\
\bottomrule
\end{tabular}

\end{table}

\paragraph{Setup}
Alpasim drives a policy inside neural reconstructions of real drives
\cite{yang2023unisim}. Open loop scores the plans of \Afive against the logged
future, without re-planning; the score is a negative displacement cost, a fixed weighted sum of the average and
final displacement errors at 3 and 5\,s, each divided by a fixed scale. The
controller is trained on \NuRecPoolScenes{} scenes and evaluated on
\NuRecEvalScenes{} held-out scenes from disjoint clips; its anchor, measured on
the training scenes, is $w{=}\NuRecAnchorW$.

\paragraph{Online RL baselines}
Discrete soft actor-critic (SAC) \cite{haarnoja2018sac,christodoulou2019discretesac}
and a group-relative (GRPO-style) policy gradient \cite{shao2024deepseekmath}
learn the same per-step weight from the same state, action set, task score and
training scenes, trained from scratch. SAC's reward is paired against the $w{=}1$
rollout, and GRPO is regularized toward the anchor policy.

\paragraph{Results}
In open loop every stage of the controller is above every fixed weight, SAC and
GRPO (Table~\ref{tab:nurec}); at its third stage it lowers the displacement
cost by \NuRecPeakGainPct\,\% against $w{=}1$. Closed loop drives the same
policy inside the reconstruction, every frame re-rendered from the vehicle's
actual pose. The controller passes the most scenes, \CLPassCtl{} of
\CLScenes{} against \CLPassWOne{} for $w{=}1$ and \CLPassAnchor{} for the
anchor, with the fewest at-fault collisions, \CLCollAFCtl{} against
\CLCollAFWOne{} for $w{=}1$ and \CLCollAFAnchor{} for the anchor
(Table~\ref{tab:closedloop}).

\subsection{Waymo End-to-End Driving}
\label{sec:waymo}

\paragraph{Benchmark and system}
The Waymo Open Dataset end-to-end driving benchmark
\cite{xu2026wode2e,waymo2026e2eleaderboard} poses vision-based planning on
long-tail scenarios. Its rater feedback score (RFS) is the mean of eleven
per-category scores of a plan against trajectories rated by humans. Of
\WaymoValFrames{} scored validation frames, \WaymoEvalFrames{} are held out for
every controller comparison and \WaymoTrainFrames{} supply training data; the
organizers score the hidden test set. \Afive \cite{nvidia2025alpamayo} decodes
its trajectory in $K{=}\DiffusionSteps$ Euler steps. It is trained on a camera
rig and a vehicle unlike Waymo's, so rule-based conversions map Waymo's camera
images into its camera geometry and its plan onto the Waymo vehicle. The VLM
and the expert stay frozen; the anchor is the default $w{=}1$.

\begin{table}[t]
\caption{\textbf{Waymo, held-out, \Afive.} RFS on the \WaymoEvalFrames{}
held-out frames, every arm with the same conversions; $\Delta$ is ours (stage
$n{=}2$) minus the row. $\ddagger$: one weight per scene, scored in a separate
evaluation of the same frames; its $\Delta$ is against the per-step controller
in that evaluation, its four stages averaged per scene (\PerStepScore{}).}
\label{tab:waymo}
\centering
\scriptsize
\setlength{\tabcolsep}{3pt}
\begin{tabular}{@{}lcl@{}}
\toprule
method & RFS & $\Delta$, ours minus row \\
\midrule
fixed $w{=}0.25$ & 8.1431 & $+0.160$ \\
fixed $w{=}0.5$ & 8.2096 & $+0.094$ \\
fixed $w{=}0.75$ & 8.2412 & $+0.062$ \\
fixed $w{=}1.0$ (default, anchor) & 8.2363 & $+0.067$ \\
fixed $w{=}1.25$ & 8.1736 & $+0.130$ \\
fixed $w{=}1.5$ & 8.1197 & $+0.183$ \\
fixed $w{=}2.0$ & 7.9734 & $+0.330$ \\
\midrule
CFG++ \cite{chung2024cfgpp} & 8.0084 & $+0.295$ \\
linear ramp & 8.2419 & $+0.061$ \\
guidance interval \cite{kynkaanniemi2024interval} & 8.2418 & $+0.061$ \\
guidance rescale \cite{lin2024rescale} & 8.2441 & $+0.059$ \\
reward fine-tuning of the expert \cite{dong2023raft} & 8.1911 & $+0.112$ \\
discrete SAC \cite{christodoulou2019discretesac} & 8.2783 & $+0.025$ \\
GRPO-style policy gradient \cite{shao2024deepseekmath} & 8.2493 & $+0.054$ \\
\midrule
scene-level $Q$, one $w$ per scene$^{\ddagger}$ & 8.1105 & $+0.0867$ \\
\midrule
\textbf{Q-adaptive guidance (ours), $n{=}2$} & \textbf{8.3032} & \\
\quad stages $n{=}1$ / $3$ / $4$ & 8.2917 / 8.2922 / 8.2780 & \\
\bottomrule
\end{tabular}

\end{table}

\begin{table}[t]
\centering
\caption{\textbf{Waymo end-to-end driving, official hidden test.} Rater feedback score (RFS) of the ten best
published methods with a reported hidden-test score, one row per paper with its best reported variant, as printed in its paper
or in the hidden-test tables of~\cite{yuan2026suv,zheng2026drivema}; four-decimal scores are rounded to three.}
\label{tab:official}
\footnotesize
\setlength{\tabcolsep}{6pt}
\begin{tabular}{@{}lc@{}}
\toprule
Method & RFS $\uparrow$ \\
\midrule
SimWAM~\cite{zhao2026simwam} & 7.84 \\
FROST-Drive~\cite{dong2026frostdrive} & 7.856 \\
MindVLA-U1 + RL~\cite{huang2026mindvlau1} & 7.87 \\
IRL-VLA~\cite{jiang2025irlvla} & 7.89 \\
Qwen-Drive-1.0-RL~\cite{zhou2026qwendrive} & 7.91 \\
SUV~\cite{yuan2026suv} & 7.94 \\
Poutine~\cite{rowe2025poutine} & 7.986 \\
RAP~\cite{feng2025rap} & 8.043 \\
NTR (ensemble)~\cite{li2026ntr} & 8.046 \\
DriveMA-4B~\cite{zheng2026drivema} & 8.079 \\
\midrule
\Afive{} + Q-adaptive guidance & \textbf{\OffQTwo} \\
\bottomrule
\end{tabular}
\end{table}

\paragraph{Held-out split}
Table~\ref{tab:waymo} compares all \OneSessFixedTotal{} fixed
weights, four adaptive-guidance methods ported onto the axis with hyperparameters swept
on the training frames, reward fine-tuning of the expert at $w{=}1$, SAC and
GRPO as in Sec.~\ref{sec:alpasim}, and a scene-level controller, a regression
that picks one weight per scene from trace and command features, fitted to each
constant weight's score on the training frames. Stages $n{=}1$ to $3$ clear every fixed
weight, adaptive-guidance method, reward fine-tuning, SAC and GRPO, and stage
$n{=}2$, the best on this split, reaches \OneSessQTwo{}: \OneSessQTwoOverFixedThreeQuarter{}
above the best fixed weight ($w{=}\OneSessBestW$) and
\OneSessQTwoOverFixedOne{} above the default. Averaged over its four stages,
the controller is \PerStepOverSceneLevel{} above the scene-level controller
($\ddagger$ in Table~\ref{tab:waymo}).

\paragraph{A rule on the late disagreement}
The \emph{late disagreement}, the distance between the latents of the constant
$w{=}1$ and $w{=}0$ rollouts entering the last Euler step, exists only after
both have run. A receding-horizon rule obtains it at decision time: at every
Euler step it rolls candidate schedules ahead and keeps $w{=}1$ unless the
reasoning's exposure on the last steps would exceed a calibrated limit.

\paragraph{Hidden test set}
We run the stage-$n{=}2$ controller, the best on the held-out split
(Table~\ref{tab:waymo}), on the hidden test set; it chooses the weight at every
Euler step. The system sets a new state of the art among published methods
(Table~\ref{tab:official}): \OffQTwo{} RFS against \OffBestPublished{} for
\OffBestPublishedName{}.

\section{Conclusion}
What matters is not whether to trust a VLA's trace but how strongly it steers.
\coccfg{} makes that a control axis inside flow matching, and the frozen,
resettable generator gives each step's weight a measured value. Q-adaptive
guidance learns from these measurements; on \Afive it outperforms every fixed weight and online
RL in Alpasim open loop, passes the most closed-loop scenes, and sets a new
state of the art among published methods on the Waymo end-to-end driving
benchmark.

{\footnotesize
\ifdefined\bibsep\setlength{\bibsep}{1pt plus 0.5pt}\fi
\bibliographystyle{ieeenat_fullname}
\bibliography{arxiv/references}
}

\end{document}